# The Monitor Model and its Misconceptions: A Clarification

Michael Carl,
Kent State University

## Abstract

Horizontal (automatic) and vertical (control) processes have been observed and reported for a long time in translation production. Schaeffer and Carl's Monitor Model integrates these two processes into one framework, assuming that priming mechanisms underlie horizontal/automatic processes, while vertical/monitoring processes implement consciously accessible control mechanisms. The Monitor Model has been criticized in various ways and several misconceptions have accumulated over the past years. In this chapter, I update the Monitor Model with additional evidence and argue that it is compatible with an enactivist approach to cognition. I address several misconceptions related to the Monitor Model.

## 1. Introduction

Over the past 40 years, Translation Process Research (TPR) has developed numerous models to explain "what goes on the mind of translators". One influential theory in this respect is the computer theory of mind (CTM). The CTM views the mind as a system of computations, a program in the brain, that manipulates symbols according to a set of rules. The program that runs in the brain could equally be implemented and executed on a machine, and thus simulate human intelligence and cognition, outside the brain. The symbols that are believed to be processed in the brain (or another computational device) are physical states that are manipulated with truth-preserving rules, regardless of their semantic content. Pereplyotchik (2016, 171) points out that in "classical" cognitive architectures, such as ACT-R, behavior is thought to emerge from production rules that fire in response to dynamically changing contents of buffers[1]. While the buffer states are thought to be mental representations (symbols) of some outside state of affairs, the procedural knowledge that is encoded in production rules is not representational. Probabilities, frequencies, rhythm of action, etc., emerge from firing rules but are non-representational in nature.

Mental representations are, according to this view, the basis on which we produce (i.e., compute) inferences, e.g., infer meaning, plan and execute actions, including translations. Accordingly, "cognitive processes can be described and explained as [a] manipulation of formal symbols in a language of thought" (Martín de León 2017, 109). Translation, under a CMT, is thus a type of mechanical theorem-proving, where "[t]ranslation was understood as a rule-guided transformation of symbols from one code into symbols of another code" (Martín de León 2017, 109). In support of this functional, computationalist view, Bechtel and Abrahamsen (2002, 191) maintain that "the time consumed by a particular cognitive process is a matter of implementation and does not inform us as to the nature of the architecture [of the mind] itself." The temporal structure of these operations is of no interest for the CTM.

---

[1] a limited set of 'variables' that are believed to exist in a cognitive module.

Criticism of the CTM took shape in the 1980s. The computationalism controversy revolved mainly around the notion of whether the mind produces and manipulates mental representations of an outside world. Perhaps not by coincidence, starting from the mid-1980s, Translation Process Research (TPR) set out to investigate "What happens in the heads of translators" (Krings 1986, see also Königs 1987) and to assess "by what observable and presumed mental processes do translators arrive at their translations?" (Jakobsen 2017, 21). A large body of data and research findings has been produced in the past three decades that documents, among other things, the role of expertise, and translation directionality, ergonomic, linguistic, and emotional factors, as well as the usage of (external) resources - such as computer assisted translation and machine translation (MT) - on the translation process. TPR focuses thereby on process related issues, such as translation duration, translation effort, and the distribution of attention. In contrast to CTM, the temporal structure of cognitive operations is of central importance in TPR.

Theories and models that reject the CTM are non-computationalist or postcognitivist; some are also non-representational. Sometimes these theories are labeled 4EA cognition (embodied, embedded, enactive, extended, and affective). Postcognitivists have proposed a large number of alternative theories (for an overview see e.g., Risku and Rogl 2020) which suggest that the body and the environment have a facilitating or even a constitutive role in cognition. According to some postcognitivists, cognition does not only take place in the brain but may be distributed in the body and/or the environment, knowledge is embodied and situated, emotions and rational thought cannot be separated in human experience, cognition and action can be direct without intermediate mental representation of an external world, which led some researchers to assume that the brain is a resonant organ rather than a representational one (Ryan and Gallagher 2020). In this view, the temporal structure associated with cognition and action is crucial for the understanding of the involved processes.

Hutto and Myin (2017), for instance, endorse a radical enactivist version of postcognitivism. They make a distinction between basic cognition and higher-order, content-involving cognition. Basic cognition, they say, is intentional but does not involve representational content[2]. Higher-order mental representations are assumed to have "special properties – e.g., truth, reference, implication – that make it [the representations] logically distinct from, and not reducible to mere covariance relations" (Hutto and Myin 2013, 67). Basic cognition is non-representational, as it does not specify representational content. However, the notion of content is "elastic enough" to include "accuracy, veridicality or some other kind of satisfaction condition" (*ibid*.).

Within an enactivist framework the question thus shifts from what counts as a representation of "real entities" to whether or to what extent do translators engage in basic or higher-order (e.g., propositional) content to produce translations. Rolla and Huffermann (2021) extend Hutto and Myin's framework with a notion of *shared know-how*. Linguistic agency, they say, requires sophisticated forms of shared know-how on which the normativity of human cognitive capacities rests. Rightness and truth are in this view relative to paradigmatic sets of actions, rather than to external entities and their mental representations. Rolla and Huffermann suggest a hierarchy of three different levels of content in which higher-order content implies the existence of lower-level content. All levels of content and all kinds of cognitive performance involve know-how with different, but stable and reproducible success conditions.

---

[2] For Chemero (2010, 77) 'non-representational' can be any state in a processing chain "produced by one part of the system, for the use by some other part of the system." This notion of basic representation does not assume any particular higher-order content or satisfaction condition. But see Rolla and Huffermann (2021).

Not everyone agrees with the notion of (non-) representationalism. Carey (2009), for instance, finds the notion of non-representationalism "puzzling." According to her, there are different types of sensory/perceptual, core and conceptual representation, which have different characteristics, independent from their "veridical" status. Perceptual representations, she says, "are iconic or analog, whereas at least some conceptual representations are stated over discrete, arbitrary symbols" (*ibid*. 8). "Representations are graded in robustness or strength, are constructed in real time, and are subject to multiple interacting influences during the processes of construction" (*ibid*. 48). Representations in "core" cognition "need not be (and often are not) veridical and therefore need not be knowledge" (*ibid*. 10). What Hutto and Myin take to be a defining criterion for representation-hood (i.e., truth conditions) is, for her, the veracity of representations. Despite those differences, researchers seem to agree, however, that there are different levels of representation and processing. And for the translation process, specifically, different processing levels have been conceptualized in the Monitor Model (Schaeffer and Carl 2013; 2015) in terms of automatic vs. monitoring processes.

In section 2, I trace the development of the Monitor Model within a historical context. I show how the Monitor Model was conceived as an approach to integrate automatism and monitoring processes in translation. Section 3 develops the concept of automatism in translation in more detail and shows how it can be considered an instance of goal-oriented basic cognition. Priming, I suggest, is an automatism that is important in translation production and a form of basic cognition that does not involve mental representation and specifications of truth conditions. In section 4, I address some misconceptions regarding TPR and the Monitor Model. I argue that TPR and the Monitor Model are not primarily focused on developing a language of thought or on studying representational or conceptual content, such as *truth, reference, or implication* of translations. Rather, TPR and the Monitor Model share many features with a non-representational, radical enactive framework, stipulating that much of translational activity can be explained in terms of basic content.

## 2. TPR and the Monitor Model

In the editorial introduction to Krings (2001) the translation process is described as follows:

> When human beings translate, they construct meanings from the sentences they read, then take this meaning and express it in another language, taking into account all of the nuances of the source and target cultures, the textual world of the text in both cultures, and their knowledge of the languages involved and the differences between them (Koby, in the editor's introduction to Krings 2001, 6-7).

Despite the fact that this statement implies the existence of representations of culture, language, and textual meaning, the main focus of Krings' work was not on representational content, but rather on the dynamics of post-editing *effort*. Using Think-Aloud (TA) protocols, his work aimed at overcoming the "complete lack of empirical, controlled observations concerning all issues associated with the post-editing process" (*ibid.* 65). He introduced the famous distinction between temporal, technical, and cognitive effort and provided numerous examples that explain his conception of translational effort, but he does not address issues of reference or truth conditions in translation.

The same holds for Lörscher (1991) who used TA protocols to investigate problem solving and translation strategies of professionals and university students. According to Lörscher, there are two kinds of translation: automatic, nonstrategic translation, and controlled strategic translation which involves problem solving. TA is indicative of strategic translation; little TA can be observed in phases of automatic language processing. Lesser TA for professional translators shows, therefore, that they have reached a higher degree of automatization. However, as a translation process researcher, Lörscher was less interested in developing a (higher-order) notion of representation, or even assessing the quality of translation. The development of a representational CTM was not primarily at stake in earlier TPR.

In the 1990s, keystroke logging and eye tracking technologies were introduced to investigate the translation process at a much finer-grained temporal resolution (below the level of a second). Based on the assessment of such logs it has been suggested that at least two concurrent processes take part in the translation process. Englund-Dimitrova (2005, 26) notes that "there are segments which are translated apparently automatically, without any problems, and other segments where the translation is slow, full of many variants and deliberations, which necessitates a problem-solving approach and the application of strategies." Based on similar observations, Tirkkonen-Condit (2005) distinguishes a default translation procedure (or automaton) and its monitoring mechanism. The default translation automaton[3] "operates on a lexical as well as syntactic level" (*ibid.* 405) and produces translation hypotheses while the "monitor supervises text production processes, and triggers disintegration of the translation activity into chunks of sequential reading and writing behavior" (Carl and Dragsted 2012, 127). Indeed, much of TPR focused on conceptualizing, measuring, and evaluating various aspects of lexical and syntactic (translation) ambiguity and its impact on effort and quality in post-editing and from-scratch translation.

Schaeffer and Carl (2013; 2015) introduced a recursive model of the translation process in which priming mechanisms are identified to activate shared lexical and syntactic representations[4]. Automated translation production is driven by priming processes in which "features of both source and target language items [are activated] which share one single cognitive representation" (2015, 21). However, shared, non-selective activation (de Groot 1997) implies distributed representations and tensor product parsing processes which do not share the structure of the sentences they parse. These non-isomorphic cognitive structures do not account for assumed systematicities that classical computationalism suggests. Priming – as modeled through the translation automaton – may be considered a form of basic cognition that establishes the bulk of translation relations. Vertical monitoring processes, in contrast, work in a monolingual mode, controlling and evaluating the output of the translation automaton: "Vertical processes access the output from the automatic default procedure recursively in both the source and the target language and monitor consistency as the context during translation production increases" (*ibid.* 38). Schaeffer and Carl maintain that "[o]nly after the output has been received and evaluated by some kind of monitor can these processes become consciously controlled" (*ibid.* 26).

The distinction between more or less automatized processing is not addressed in "classical" computationalism. Cummins (1989, 20) speculates that "perhaps consciousness isn't essential to the mind in the way that cognition is." Cognition, in his view, is the processing of basic mental symbols in an internal language of thought (LOT) with a context-free grammar. But Fodor (1983) himself is skeptical that understanding conscious thought (i.e., operations in the central processor) can be possible at all. Fodor's (1983) "First Law of the Nonexistence of Cognitive Science" states that conscious processing is non-modular (i.e., context sensitive), and thus computationally intractable; accordingly, it will never be possible, to explain higher-order cognitive processes because they cannot be understood and implemented in a LOT. But the assumption of non-selective, distributed activation in modular cognitive domains (e.g., de Groot 1997) and the resulting non-isomorphism of basic mental representations puts into question the LOT hypothesis altogether. It is unclear whether there is any clear line between rules and representation in the mind. Processes based on the vector or tensor product (e.g., Bechtel and Abrahamsen 2002, 165 ff) suggest that representations are mapped through processing layers associated with related information, rather than through rule-based systems.

---

[3] Also labeled literal translation automaton or literal translation procedure
[4] Primed representations are basic representations: they carry basic content; they do not specify propositional content of truth, implication, or higher-order accuracy conditions.

The Monitor Model was first and foremost conceived to "investigate automated processing during translation" and to point out ways how to integrate automatic processes with higher-order controlled processes[5]. Priming relates to the observation that the processing speed of a new stimulus depends to a large extent on the relatedness of a preceding stimulus. As stimuli are provided through the external environment, priming (and thus the working of the automaton) inherently depends on environmental features that determine translational activity. Based on this insight, Carl (2021a, 2022) argues that priming processes are part of perception-action loops in translation that are the basis for tight coupling of translators and their (technological) translation environments. Priming, in this view, can be seen as a dynamic process evolving through state space, rather than the rule-based processing of mental representations.

Translation effort and the assumed interaction between horizontal and vertical translation processes have mainly been studied on a rather abstract level through analyzing accumulated typing duration and gazing patterns. Sequences of uninterrupted typing are taken to represent horizontal/priming processes (Alves and Vale 2011) whereas revision, pauses and hesitations are taken to indicate more effortful vertical monitoring processes. A case in point is pause analysis. Pause analysis has been one of the main topics of TPR (O'Brien 2006; Alves and Vale 2009; Timarová et al. 2011; Carl and Dragsted 2012; Lacruz et al. 2012; Kumpulainen 2015). Pause analysis examines the lag of time between successive keystrokes during text production and can be combined with gazing data (or other behavioral measures). However, rather than investigating the dynamic interaction of horizontal and vertical processes, the preferred method has been to average production duration over segments of texts, typically a sentence. Lacruz et al. (2012), suggest a number of measures. For instance, the pause-to-word ratio which measures the amount of fluent typing, and thus the ratio of assumed priming and monitoring processes. They suggest that pauses relevant for assessing cognitive effort (i.e., vertical processes) should be longer than 300ms, as inter-keystroke delays of up to 300ms could be due to problems of motor control.

While automated priming processes work fast and with restricted memory resources, the recursive Monitor Model does not exclude vertical control processes from having access to a (potentially) large amounts of cognitive resources, including representations of context, truth, implication, equivalence, etc. However, the main ambition of TPR has been to find out what variables determine behavior and how to explain notions such as translation effort, expertise, creativity, etc. TPR is mainly concerned with aspects of cognition analyzing the relation between process and product. Aspects of representational content are not primarily the focus of *process* research.

---

[5] The Monitor Model does not make any assumptions on the modularity of monitoring processes.

### 3. The translation automaton

The translation automaton within the Monitor Model simulates translation priming processes. Priming processes allow the mind and the environment to integrate in an online translation process. In order to illustrate the notion of *online cognition*, Gallagher (2017) and Chemero 2010), among others, provide an example in which the fielder in a baseball game follows the trajectory of the ball by keeping it visually stationary on the retina. Running after the ball while looking firmly at the ball gets him to the catching spot without the "need to compute mental representations of the ball, its speed, its trajectory, and so on" (Gallagher 2017, 14). Similarly, during phases of automated translation, the translator's eyes scan the source text at a steady pace, while the fingers type out translations with an almost constant eye-key span (cf. Carl and Dragsted 2012). Similarly to the fielder, who arrives at the catching spot without computing representational content of the situation[6], an (experienced) translator may arrive at the end of the clause or sentence with little (or only minimal) local disturbances and adjustment, and without computing the representational content, assessing the truth, reference, or implications of the translated piece of text[7]. Such patterns of online translation production specify success conditions (e.g., arrive at the end of the sentence) of basic content. Basic cognition stipulates that "non-representational and non-conceptual content ... emerges through the open-ended ongoing interactions between agents and their environment." (Rolla and Huffmann 2021, 2).

Translation process data supplies ample evidence for such conclusions and supportive evidence for the existence of automated (subliminal) translation processes, among other things, also come from TPR. García (2019, 165) posits that "translation equivalents can be accessed unconsciously." Hvelplund (2016) reports that automated translation processing is more frequently encountered in expert translators. Similarly, Schaeffer et al. (2016) found that less translation ambiguous words have an effect on automatic processes, as indicated by first fixation duration. These are indicators of basic content during horizontal translation processes in which the mind and environment are directly coupled in affordances of perception-action loops. The effect of frequencies on these processes undermines the computationalist assumption of truth and correctness criteria that are supposedly associated to processed symbols.

The neural perspective also gives evidence for automatized translation processes. For instance, García and Muñoz (2020) report that translation production is possible without comprehension (i.e. inferencing), and that lexical processing may be disconnected from syntactic processing. They also argue that there are different processes depending on translation directionality, form-based and conceptually mediated translation, translation of low and high frequency words, etc. García and Muñoz (2020, 242) explain that cognitive mechanisms can become "functionally autonomous, in the sense that one [processing module] can become impaired while the other(s) remain(s) partially or fully functional."

---

[6] Such as calculating the overall trajectory of the ball, or the time of arrival and his own speed, etc.
[7] Such processes can be modelled by a finite state automaton, for instance a Hidden Markov Model (e.g., Carl 2012). This idea assumes that the observed patterns are produced by hidden processes that can be described as sequences of interconnected states. The hidden states consume a segment of input (e.g., read a source segment) and generate (emit) output (e.g., typing of a translation). Transitions between the states may depend on numerous parameters and they can be probabilistic, but the consumed resources and inference mechanisms are assumed to be limited, which makes an automaton quick, as compared to more sophisticated reasoning systems. Note that a notion of automaton does not specify the type of information or content processed; however, they more likely pertain to a notion of basic content.

These findings have not only been shown in lesion studies of brain-damaged bilinguals but also in healthy people. It might be possible that experienced translators acquire skills to functionally disconnect brain regions, so as to optimize translation production, and that they (re-)activate or integrate those processes as they are needed. This would allow them to translate (at least partially) without a need to compute and evaluate mental representations and/or integrate the meaning of entire segments. Hutto and Myin (2017, 26) explain that the development of such skills and expertise "depends on structural changes inside the organism," the ability to dynamically distinguish between and activate more fine-grained different processes, experiences, and situations and to react accordingly, but not necessarily to build up more sophisticated representations.

Further, Dehaene (2014, 52ff) provides numerous examples that illustrate how emotion and meaningful action can be detached from awareness, consciousness and thus (higher-order) representation. In masked priming experiments, he shows not only that semantic processing can be triggered unconsciously, but also that items of "extremely automatic and overlearned" sequences – such as vision and speech – can be unconsciously assembled (binding) and integrated even across different modalities, where the "brain can unconsciously process syntax and meaning of well-formed word phrases" (*ibid.* 73). Dehaene supposes that "routine bindings would be those that are coded by dedicated neurons committed to specific combinations of sensory inputs. Nonroutine inputs, by contrast, are those that require the *de novo* creation of unforeseen combinations – and they may be mediated by a more conscious state of brain synchrony" (*ibid*. 63).

In the same line, Carl (2021a) suggests that the translation automaton may be considered a loop of action-perception in which "we see things [i.e., words and segments of the source] in terms of what we can do with them" (Gallagher 2017, 41). According to this view, cognitive processes are not just in the head but involve bodily and environmental factors: "sensory–motor contingencies and environmental affordances take over the work that had been attributed to neural computations and mental representations" (*ibid*. 7).

In this view, at times the monitor intervenes, monitoring processes disintegrate loops of concurrent perception-action into successive reading and typing activities (Carl and Dragsted 2012). In his embodied version of the Monitor Model, Robinson (2020, 138) adopts the notation of Kahneman (2013) and suggests that the translation automaton (i.e., System-1) is responsible for translational "norm-formation [which] relies at a very deep level on affect management". Robinson suggests that monitoring processes (i.e., System-2) are triggered through enhanced awareness of affects: "*feeling* affective body states does significantly guide the translation from System-1 automaton to System-2 emotional intelligence" (*ibid.* 139). Kahneman (2013) posits that System-2 may be lazy, which he attributes to a "flaw in the reflective mind, a failure of rationality" or it may be engaged, "more alert, more intellectually active, less willing to be satisfied with superficially attractive answers" (Kahneman 2013, 49), and thus less susceptible to cognitive errors. "System 2 is in charge of doubting and unbelieving, but System 2 is sometimes busy, and often lazy. System 1 produces most of the decisions but "favors uncritical acceptance of suggestions and exaggeration of the likelihood of extreme and improbable events" (Kahneman 2013, 61). Kahneman also points out that "[t]he combination of a coherence seeking System 1 with a lazy System 2 implies that System 2 will endorse many intuitive beliefs, which closely reflect the impressions generated by System 1" (*ibid*. 96).

There is an ongoing discussion as to how this (apparent) discontinuity between automatism/conscious processing, System-1/System-2 or basic and higher-order cognition can be consolidated from a philosophical perspective (e.g., Gallagher 2017; Hutto and Myin 2017; Rolla and Huffermann 2021) or with respect to neural processing. Dehaene suggests an exponential increase of brain activity when conscious processing takes over. System-2 is more likely linked to conscious processing, and the minimization of cognitive effort may explain the preferred laziness of System 2. However, depending on the theoretical assumptions (e.g., first-order vs. higher-order theories) this does not necessarily imply that all conscious brain states can be identified with higher-order representational content.

Whatever the mechanisms are that trigger monitoring processes, it might be hypothesized that representational content could be generated as monitoring processes intervene, triggering rational thought and inferencing over mental representations which lead to successive translational activity. Thus, Hutto and Myin (2017, 101) note that "an organism's skillful engagements with the world are best understood in embodied, enactive, and nonrepresentational ways" while representational content may be generated when the agent encounters perturbations during ongoing activity (*ibid*. 77). But even in challenging translation situations, the eyes usually remain firmly on the text (Carl and Dragsted 2012), actively searching and sampling words that fit expectations or resolve uncertainty (cf. Wei 2021). Thus, even in those instances of search and monitoring, a decoupling of the stimulus (i.e., the text) does not necessarily take place – which would be one of the indicators for (higher-order) representation. Gallagher (2017, 93) supposes that there may be "some mechanism that may (or may not) operate in a representational way in a non-action (so-called higher-order cognitive) context," but it is doubtful whether this also "means that it is necessarily operating in a representational way in a perception-action context." As Gallagher (2017, 94) points out, "the environment itself does some causal work, and it does so in a way that undermines the notion of decouplability and eliminates the need for [higher-order] representations."

The Monitor Model suggests that meaning hypotheses "are constructed to the extent and at the moment they are needed to continue the task at hand." (Carl and Dragsted, 2012, 144) which implies a fluid and dynamic interaction between the translators and their textual/technological environment. Similarly, Hutto and Myin, (2013, 82) assume that a "great bulk of world-directed, action-guiding cognition exhibits intentional directedness that is not contentful ... It is possible that even sophisticated forms of human visual perceiving are not essentially contentful or representational", and at least some findings of TPR show the same may also be true for translation. Especially experienced translators often type out translations after reading only two or three words ahead in the source, thus without possibly knowing the content or structure of the source sentence (cf. the 'head-starter' in Carl et al. 2011). These translators cannot possibly have an internal representation of the sentence proposition, its truth values, reference, or implications – more likely translations emerge here merely as primed covariance pattern[8]. While the Monitor Model and the finding that "we act before we think" is compatible with a radical enactivism, it is "an outright denial of the CIC [Content Involving Cognition] thesis that 'we must think in order to act'" (Hutto and Myin, 2013, 12). It is precisely this non-representationalism that is rejected by some CT proponents as it allegedly "make[s] us partial cyborgs who do not even know what they are processing until something goes wrong" (Muñoz 2016, 16).

---

[8] This conclusion is also implied in frequency-based models such as Halverson's salience and gravitational pull hypotheses and in models of bilingual procession, see e.g., Halverson (2015). It suggests that translators take risks, engaging in formulation of translations without complete knowledge.

## 4. Misconceptions of the Monitor Model

The Monitor Model suggests that much of translational activity can be explained in terms of non-representational basic cognition, but at the same time it does not exclude representational monitoring processes from interfering with automated processes. Proponents of Cognitive Translatology (CT) (Muñoz (2016); Martín de León (2017); Muñoz and Rojo López (2018), Muñoz and Martín de León (2020); and others) seem to reject the Monitor Model based on objections which I will address in this section.

### 4.1. Translation and priming

Muñoz (2016) was perhaps the first to critically analyze Schaeffer and Carl's conception of the Monitor Model, labelling it a "computationalist account," due to its notion of translation automaton. Muñoz rejects the notion of "automaton"[9] as "the way translation unfolds is unconvincing" and he wonders "why and how we should have it" (*ibid.* 15). He acknowledges that priming occurs "outside of conscious awareness" (*ibid.* 6) and at the same time maintains that meaning is the "transient result of the interaction of … referential, combinatorial, emotional-affective, and abstract information." However, he does not address in his own model how priming and (higher-order) inferencing – two apparently contrary processes – could be reconciled. It is not clear whether the Monitor Model and its translation automaton are rejected because it is assumed that priming plays a (crucial) role in translation or because of the view that an automaton is inappropriate for modeling priming effects. He lists several requirements which he claims cannot be modeled by an automaton, including expertise, experience, time and fatigue, frequency effects of translation entrenchment and saliency, random choices, and emotional and affective states.

However, given the appropriate configuration, an automaton can certainly simulate all of the requirements listed above. Under certain restrictions, even neural networks can be considered a type of finite-state automaton, which have been shown to be sufficiently powerful for simulating priming effects (e.g., Hartsuiker et al. 2004 and 2008, van Gompel 2013, Eddington and Tokowicz 2013, Frank 2021). In contrast to monitoring processes, the translation automaton is assumed to be an instance of basic cognition with basic content and access to restricted resources.

Perhaps even more problematic is Muñoz' notion of computationalism as a way to describe the Monitor model. As he points out, one of the crucial properties of priming is that no (higher-order) mental representations are produced. Translation production without higher-order representation is, however, the main function of the automaton within the Monitor Model. By situating the translation automaton within a "computationalist account," Muñoz must thus either reconsider his notion that a "computational theory of mind is often considered a representational theory" (Muñoz 2016) or find a way to reconcile translational priming effects within CT and his assumption that priming does not play a role in translation, i.e., "the divide between lower and higher functions is not that important" (*ibid.*). The former course of action would lead to a pancomputationalist (or perhaps a mechanistic) position, which does not seem to be consistent with his overall perspective. In the latter case his argument would first of all be geared against the role of priming in translation (see below) and not so much against the notion of automaton. In any case, we argue that a generic term '"computationalist account" is a misnomer for the Monitor Model.

---

[9] In the same paper, Muñoz (2016) describes a translation automaton as follows: "We only have one vast memory store, where language form representations work as pointers to recurring patterns of neural processing … In order to navigate through this store and perform different operations—what we call access and retrieval, working memory, etc. … [cognitive operations] link details of language units, such as aspects of shape, register, grammatical features, and language membership" (Muñoz 2016, 15). See also footnote 19

### 4.2. Translation as a dynamic process

Following Muñoz, also Halverson (2019, 204) asserts that "Schaeffer and Carl represent the classical computational approach to cognition, which concerns itself with problem solving and algorithmic processes". She maintains that she follows a non-computational approach and does not consider her own work to be affiliated with any "classical computational approach", thus rejecting the Monitor Model and related work of Schaeffer and Carl. However, she is not explicit about what she means by computationalism (or non-computationalism)[10]. It remains unclear what "algorithm" and "problem solving"[11] might mean to her and how this relates to a CTM, and her own research. She does not seem to endorse pancomputationalism or a mechanistic approach to computation either[12]. If we assume an algorithm to be an efficient way to solve a problem, should we infer that all algorithmic (i.e., systematic or efficient) descriptions of problem-solving processes are part of a CTM?

To support her argument, Halverson (2019) identifies two papers supposedly following a "classical computational approach." The first (Schaeffer and Carl 2017) is an analysis of translation process data which develops – among other things – a perception-action model of translation. It is an attempt to address and structure the internal dynamics of translation units. As defined by Alves and Vale (2009) translation units are minimum cycles of reading and typing activities during translation production. By analyzing traces of keystrokes and gaze behavior, Schaeffer and Carl (2017, 153) show that "reading the ST or the TT and writing constitute minimal and coherent problem identification and solution cycles." The shape of these cycles (TUs) is highly predictable through the complexity of textual material translated[13]. The analysis they present describes translation processes as a dynamic system, which is – according to the definitions given in this article – fundamentally and substantially different from a "classical computational approach."

In the second publication, Schaeffer et al. (2018) investigate similarities of translation ambiguities across different languages (see also Ogawa et al. 2021) and their (assumed) associated translational effort. Again, this work focuses on conceptualizing dynamic aspects of translational variation, rather than on representational content or rule-based symbolic inference as a classical computational approach would imply. By describing the dynamics of variation in translation, this work differs fundamentally from classical CTM approaches.

---

[10] Halverson endorses a cognitive linguistics (CL) approach to translational cognition, which relies on a higher-order notion of mental representational content in which "meaning is equated with conceptualization" (Langacker, 1986), and where conceptualization is (presumably) a precursor for translation. However, with the assumption that those representations are anchored in the body (e.g., through metaphors), and that CL rejects the independence of conventional linguistic strata (such as morphology, syntax, semantics, etc.), it is less functional than a 'classical' computational approach. CL is, therefore, said to constitute a representational, but non-computational approach to cognition.

[11] One can distinguish *heuristics* from *algorithms* in which algorithms are infallible while heuristics provide approximate solutions. In another dichotomy, algorithm could relate to know-that (e.g., symbolic step-by-step) as opposed to know-how (e.g., jump to conclusions based on intuitions, or priming, etc.).

[12] Even though her terms "gravitational pull" and "magnetism" imply the existence of physics-like laws in translational cognition. They stipulate a pancomputationalist view on cognition and a high degree of systematicity and efficiency. See also Carl et al. (2019) for a further development of those ideas.

[13] Hvelplund (2016) reports a similar investigation on a subset of the data that Schaeffer and Carl (2017) used.

### 4.3. Comprehension, transfer, and production

In another article, Halverson (2020, 40) describes the work of Schaffer and Carl as a "cognitivist/computational, information processing approach to Cognitive Translation Studies," which she suggests is different from cognitive translatology (CT) because the work of Schaffer and Carl is typical of a research tradition that considers translational cognition as "the result of 'computational' procedures that operate on representational structures in the mind"[14] and in which the "translation process is described as consisting of the stages of comprehension, transfer and production."

As discussed above, Schaffer and Carl (2013/2015) suggest that monitoring processes may (or may not) operate with a notion of higher-order representation. But this point was not the (main) focus of the Schaffer and Carl (2013/2015). In my view, monitoring processes could also be conceived of to function similarly to what is suggested in some CT models - such as Cognitive Linguistics (CL) - or, indeed preferably, in a non-representational manner.

However, it is in particular the translation automaton, i.e., translation without higher-order representation, that has produced concern and misunderstanding among some CT scholars. Schaeffer and Carl (2015,24) explicitly state that during horizontal translation "it is not necessary to fully comprehend either source or target text", which some CT scholars reject. Cognitive Linguistics (CL) (see footnote 10), for instance, stipulates that "the whole frame is required to interpret the meaning of one of its elements" (Fillmore 2000, 373, according to Serbina 2015, 69). If this is also an assumed requirement for translation, it would rather suggest that CL proponents should endorse the comprehension-transfer-production paradigm. In my view, it is, however, doubtful that priming processes can ensure all frames to be activated. The Monitor Model suggests that a full understanding (i.e., the whole frame is activated) may not always be necessary for translation. However, this does not preclude the CL framework from being instrumental in explaining monitoring processes, as for instance suggested in (Czulo et al. 2019).

In addition, Schaeffer and Carl have argued, as others have done, that translation priming involves simultaneous activation of the source and target languages, which cannot be disentangled. Carl and Dragsted (2012, 127) proposed a version of "the Monitor Model in which comprehension and production are processed in parallel by the default procedure." From the beginning, then, the Monitor Model was conceived to implement concurrent horizontal/automatized and vertical/controlling translation processes, and explicitly rejects a stratificational "comprehension, transfer and production" architecture.

### 4.4. Default translation

In an attempt to promote "opportunities for epistemic pluralism," Marín (2019, 171) claims that "Schaeffer and Carl's construct of LITERAL TRANSLATION (2014) is built on computational assumptions that are incompatible with the cognitive translatology assumptions informing Halverson's DEFAULT TRANSLATION (2015)" (emph. in the original). Marín does not specify what he means by *computational assumptions* and why he considers them to be incompatible with the assumptions of CT.

---

[14] CT proponents, including Halverson, do not make a distinction between basic and higher-order content. If higher-order representations are here assumed, then this statement does not apply to the translation automaton and thus does not describe the Monitor Model correctly. If basic representational structures (correlational content) are meant, then all cognitive processes are representational, including those assumed by CT.

However, Marín refers to Halverson (2015), which clarifies that *literal translation* refers to an "interlingual or intertextual correspondence", while *default translation* is an "unconstrained and immediate production mode." In my understanding, this view matches precisely how these terms have been used in the work of Schaeffer, Carl, and Dragsted. It is this notion and the observation of "default translations" that gave rise to the Monitor (Carl and Dragsted 2012), where default translations are assumed to be undisturbed first translational responses, produced by priming processes. Marín does not mention this origin, but the notion seems entirely consistent with Halverson's (2015, 22) definition of default translation, as "some cognitive pathways …[that] are more likely to be taken in translation". Even though *compatibility* is a central term in Marín's work on *epistemic plurality,* he does not specify why Halverson's notion of default would be incompatible with Carl, Dragsted, and Schaeffer's notion of the same term.

In contrast to *default translations*, the notion of *literal translation* as conceptualized in Schaeffer and Carl (2014) (see also Carl 2021b) relies on a comparison of alternative translations as observed in the *final translation product*. Halverson notes that "there is a principled relationship between the idea of default translation and the patterns reflected in aggregate production data" (Halverson 2019, 191), which has also been the assumption and findings in much of TPR. Several studies show a relation between properties of the first translational response (e.g., default translation) and properties of the final translation product (e.g., its literality.) However, there seems to be no incompatibility here either between the notion of Schaeffer and Carl, and that of Halverson. The Monitor Model is well suited to explain these observed relations between default translations and literal translations (Schaeffer and Carl 2013; 2015; Hansen-Schirra et al. 2017; Carl 2021c). While it explains why default translations tend to be more literal than revised translations, this does not constitute an incompatibility.

### 4.5. Concurrent processing

In the same article, Marín (2019) states that the construct of *translation competence* "is compatible with computationalist translatology" (*ibid.* 178), but again, does not specify what "computationalist translatology" might be. He explains that the PACTE competence model is inconsistent with CT (i.e., cognitive translatology), as PACTE assumes that "translation processing is based on a Monitor Model and therefore that processing is sequential." No explanation is given regarding how the Monitor Model is supposed to process sequentially and why sequential processing would be incompatible with CT - unless, perhaps, it is a corollary of Marín's (2017, 75) assumption that the Monitor Model works in discrete steps, in "alternate automatic and problem-solving processing modes."

However, while the Monitor Model assumes that monitoring "processes access the output from the automatic default procedure,"[15] Carl and Dragsted (2012, 127) propose a "monitor model in which comprehension and production are processed in parallel by the default procedure." Further, Schaeffer and Carl (2015, 36) suggest that "concurrent reading and writing during translation are indicative of automatic processes and shared representations [i.e., in the basic sense]," which also implies concurrent automatic/priming and monitoring. While behavioral data is necessarily sequential (e.g., there is typing and pausing), this does not imply that the assumed priming and monitoring processes also run sequentially. They can well be activated concurrently, but show their results in a temporal order. Parallel (or concurrent) processing is also assumed in Kahneman's system 1 and system 2, which roughly correspond to the translation automaton and the monitor, as mentioned above. Note that this is different from a discussion as to whether alternative possible translations (or interpretations) of one expression are processed in parallel or sequentially. Psycholinguistic research on garden-path sentences suggests that the monitor can access only one reading at a time (cf. van Gompel, 2013), while the non-selective activation during priming processes implies concurrent activation of all possible readings, even if to a different degree.

---

[15] This view is also suggested by the Global Neural Workspace Hypothesis (Dehaene 2014) which postulates that signals of specialized cortical processors must be strong enough to trigger conscious awareness.

Another misunderstanding regarding the Monitor model has to do with the question of whether cognition is distributed. For instance, Marín (2017, 75) also asserts that the Monitor Model is inconsistent with CT since "Muñoz's model is deeply rooted in the cognitive research tradition, which contends that processing is distributed and parallel." However, he does not point out which cognitive processes are supposed to be distributed or parallel and how this is different in CT and the Monitor Model. But it appears from their own description, (Muñoz 2016; Muñoz and Rojo López 2018; Muñoz and Martín de León 2020), that CT processes seem to be organized rather sequentially as the following indicates:

> Language units *first* activate representations in our minds, corresponding to their visual or auditory characteristics, *perhaps followed* by the activation of amodal (mainly, lexical) representations. *Then* episodic and semantic knowledge—information related to our previous experience of such units and what they mean, and what we have abstracted away as constant in them—may become activated.
>
> … the activated episodic and semantic information *may then* be integrated into the representation of the situation prompted by the text or the speech.
>
> Schemas and frames structure the knowledge in our minds, and meaning *emerges* as an inferential process, where … meaning is an online process *resulting from* the interaction between schematic, ad-hoc knowledge structures and further cognitive or construal operations. (Muñoz and Rojo López 2018:62-68, my italics)

Muñoz and Rojo López lay out here a considerable representational commitment[16]. One can, of course, assume that all these different stratificational processes run in parallel – one can hear (or read) a word while still processing a previous one, and the different processes can be distributed – lexical processing may happen in another area of the brain than syntactic processing – but if this is the case within the CT framework, then it could also be the case for the reading/writing, automatic and monitoring processes as conceived in the Monitor Model (see section 2).

---

[16] This description is somewhat similar to ideas of rule-based/unification-based MT in the late 1980s, such as Eurotra (e.g., https://en.wikipedia.org/wiki/Eurotra, Durand et al. 1991) or CAT2, in which I worked for some time, (http://www.iai-sb.com/docs/getting_started.pdf). Eurotra tried over 15 years (1978–1992) to implement a similar CT automaton into a computer program, with moderate success. Eurotra assumed six steps 1) Text normalization; 2) Morphological structure; 3) Constituent structure (frames, schemas); 4) Relational structure; 5) Interface (i.e., "meaning") structure; 6) Transfer into the TL and then a reversal through steps 5-1 for TL generation. Muñoz, however, rejects the possibility that his CT version could be implemented in a computer program, and thus undermines an important possibility for falsification. As in the early Eurotra implementation, Muñoz' description also does not specify how translations are supposed to be generated (i.e., selected) from heavily underspecified abstracted interlingual 'meaning' representations. In the Eurotra context, this lack (or maybe impossibility) of analytically formalizing translation selection skills turned out to be one of the biggest conceptual shortcomings, and arguably a reason for the unprecedented rise of statistical MT. In contrast to the Eurotra experience, Muñoz may not be able to generate similar insights, as he rejects possibilities for empirical implementation and validation (see footnote 20).

It is unclear, moreover, what makes the CT model particularly 4EA related. The suggested architecture does not appear fundamentally different from some rule-based MT architectures (see footnote 16 on the Eurotra Architecture). It is basically a continuation of the "mentalistic enterprise" of "pushing the world inside the mind" with a "very narrow-minded conception of embodiment" (Gallagher 2017,5). However, Marín accurately points out that Muñoz' and coworkers' CT model is "deeply rooted in the cognitive research tradition," which is computationalism.

## 5. Conclusion

This chapter addresses a recent discussion on computationalism and representation in Cognitive Translation Studies (CTS) and attempts to rectify some misconceptions with respect to the Monitor Model. I revisit and update the Monitor Model in the light of Hutto and Myin's (2017) notion of enactivism and basic and higher-order cognition: higher-order cognition implies conceptual content, truth, and correctness conditions of the represented items, while basic cognition builds on basic content of internal states that carry correlational information of some other states or structures (Rolla and Huffermann 2021). I identify priming processes to be instantiations of basic, non-representational cognition (a primed response is correlated with a stimulus) while monitoring processes may (also) imply a higher-order notion of mental representation (e.g., evaluation of correctness and truth).[17] The Monitor Model stipulates that much of human translational activity can be explained through basic cognition.

The chapter addresses several misconceptions of the Monitor Model. It discusses misconceived interpretations of priming and default translation, as well as stratificational, concurrent and dynamic processing. Some of these misconceptions focus on the notion of priming and the role of automatism in translation. Muñoz (2017, 564) rejects the notion of automatisms altogether ("they make us cyborgs") while the Monitor Model assumes that basic, automated processes make use of fundamentally different resources and processing routes than the higher-order monitoring ones. The Monitor Model has been rejected as a *computational account* without consideration what computational could actually mean. The term *non-computational* has been used to express a position that rejects the computational paradigm, but has been in many instances unclear which notion of "computation" is actually rejected.

One of the pillars of CT (Muñoz 2016; Martín de León 2017; Muñoz and Rojo López 2018; Muñoz and Martín de León 2020) is the assumption that "meaning is not out in language, but within our heads"[18] for

---

[17] Yang and Li (2021) assume that monitoring processes are mostly automatic and subconscious, they "can be based either on declarative/explicit knowledge or on procedural/implicit knowledge." They point out that "Expert translators may be able to interrupt the literal [i.e., automatized] translation process without even employing conscious effort, while for fledgling translators, it is most likely that conscious monitoring skills are frequently called upon." (Yang and Li 2021, 122).

[18] This view is not uncontroversial. Putnam (1973, 710) argues that "difference in extension is *ipso facto* a difference in meaning for natural-kind words" since otherwise, "'elm' in my idiolect [may have] the same meaning as 'beech' in your idiolect, although they refer to different trees", which is certainly not desirable in a translation context. Instead, Putnam requests us to "give up the doctrine that meanings are concepts, or, indeed, mental entities of any kind."

"as soon as we listen or read anything, meaning springs to our minds like a reflex."[19] (Muñoz 2016, 5) Muñoz and Rojo López (2018) suggests that these meaning reflexes are (somehow) integrated with a notion of meaning-as-inference in which cognitive operators build "mental representations [which serve] as dynamic internal support to meaning construction and to translation and interpreting processes." (Martín de León 2017, 121)

A conception of the translation process, rejecting the role of automatisms while assuming only the truth-conditional role and a higher-order concept of representations, endorses – in fact – the essential assumptions of "classical computationalism." In its formulation, CT, just as classical CTM, follows the Enlightenment's ideas of individualism (pushing the world inside the mind), and reason (meaning as rational inference) based on a higher-order notion of representational content (translating as transfer of truth). As Marín (2017) points out, CT, in this view, is firmly anchored in the computational, rationalist tradition of the enlightenment project, with surprising , but consistent conclusions[20].

The Monitor Model, in contrast, is compatible with the assumption that priming processes build the foundation of representationally unmediated brain-body-environment coupling in which interactions with the environment have a constitutive role on our thinking and meaning construction (Carl 2022). This assumption stipulates that we may have "no guaranteed privileged access to deeper facts that fix the meanings of our thoughts" over which "we, the very thinkers of those thoughts, have no special authority" (Dennett 2013,173). This undermines individualism, skepticism, and rationality, thus challenging the enlightenment project and "classical computationalism." It subverts the presumption of the rational human mind as being at the center of the (translating) world, which may be one of the reasons why the Monitor Model may cause discomfort to some.

### Acknowledgements

I would like to thank Dr. Isabel Lacruz and an anonymous reviewer for the very many helpful comments and suggestions to refine my arguments. The chapter would not be what it is without their input. All errors and misrepresentations are mine.

---

[19] Muñoz (2016) rejects the notion of a translation automaton but curiously introduces a notion of 'meaning-as-reflex' in his own model. An automaton can be probabilistic and non-deterministic; a reflex is a restricted kind of automaton which is usually deterministic or conditioned on very few parameters. A reflex uses dedicated neural pathways for faster processing, bypassing the brain altogether, a very controversial assumption to explain translational behavior.

[20] Muñoz thinks that MT systems "cheat" as they "pre-assign meaning equivalence between language units when meaning is not there, but only in our minds" (Muñoz Martín and Martín de León 2020, 67). Consequently, "a machine translation is technically not a translation ([since] there is no [human] processing of meaning involved)." (Xiao and Muñoz, 2020) For an alternative view see footnote 18.